\documentclass[conference]{IEEEtran}
\usepackage{balance}
\IEEEoverridecommandlockouts 
\usepackage{graphicx}
\usepackage{algorithm}
\usepackage{algpseudocode}
\usepackage{listings}
\usepackage[OT4,T1]{fontenc}
\usepackage[cmex10]{amsmath}
\usepackage{url}
\usepackage{multirow}
\usepackage{cite}
\usepackage[symbol]{footmisc}

\title{Neuro-Endo-Trainer-Online Assessment System (NET-OAS) for Neuro-Endoscopic Skills Training}

\author{\IEEEauthorblockN{Vinkle Srivastav\IEEEauthorrefmark{1},
Britty Baby\IEEEauthorrefmark{1},
Ramandeep Singh\IEEEauthorrefmark{2}, 
Prem Kalra\IEEEauthorrefmark{1} and
Ashish Suri\IEEEauthorrefmark{3}}
\IEEEauthorblockA{\IEEEauthorrefmark{1}Amar Nath and Shashi Khosla School of IT\\
Indian Institute of Technology Delhi, Hauz Khas,
Email: vinkle.kumar@gmail.com}
\IEEEauthorblockA{\IEEEauthorrefmark{2}Center for Biomedical Engineering\\ Indian Institute of Technology Delhi, Hauz Khas}
\IEEEauthorblockA{\IEEEauthorrefmark{3}Department of Neurosurgery,\\
All India Institute of Medical Sciences, New Delhi}}

\begin{document}

\maketitle 
\begin{abstract}
Neuro-endoscopy is a challenging minimally invasive neurosurgery that requires surgical skills to be acquired using training methods different from the existing apprenticeship model. There are various training systems developed for imparting fundamental technical skills in laparoscopy where as limited systems for neuro-endoscopy. Neuro-Endo-Trainer was a box-trainer developed for endo-nasal transsphenoidal surgical skills training with video based offline evaluation system. The objective of the current study was to develop a modified version (Neuro-Endo-Trainer-Online Assessment System (NET-OAS)) by providing a stand-alone system with online evaluation and real-time feedback. The validation study on a group of 15 novice participants shows the improvement in the technical skills for handling the neuro-endoscope and the tool while performing pick and place activity.     
\end{abstract}
\footnotetext{Published at Federated Conference on Computer Science and Information Systems - FedCSIS 2017.}

\section{Introduction}
\IEEEoverridecommandlockouts\IEEEPARstart{M}{inimally} invasive neurosurgical procedures have gained the popularity in recent years due to the reduction in postoperative recovery time, morbidity, hospitalization time and cost of patient care \cite{c1}. It provides the neurosurgeon with a better visualization method of the complex surgical site with reduced damage to the intricate anatomy of the brain. Neuro-endoscopy is a minimally invasive neurosurgical procedure that uses an endoscope image projected on the 2-dimensional display to access the interior deep structures. The margin of error is minimal and the existing apprenticeship based method of training is not suitable. It requires training for eye-hand coordination, depth perception, and bimanual dexterity. The simulation-based training outside the operating room is getting wide acceptance due to the provision of repeated practice, objective evaluation, real-time feedback and staged development of skills without the supervision of an expert surgeon \cite{c4}.\par
Simulation-based training in neuro-endoscopy varies from low-fidelity natural simulations, box trainers, part-task trainers, to intermediate-fidelity synthetic simulators, virtual reality simulators and high-fidelity cadavers and animal models. The box-trainers or part-task trainers are designed to impart training for fundamental technical skills of instrument handling and eye-hand coordination. The synthetic simulators and virtual reality trainers provide training for anatomy and procedures but give limited haptic feedback. The high-fidelity simulations on cadavers and animals provide training for anatomy and procedures along with haptic feedback and realism \cite{c6, c7, c8, c9, c11}. \par 
The evaluation of the surgical activity on the various simulation systems is platform-specific. The assessment methods can be based on direct observation, error metric of the task, sensor-based evaluation of the motion and video-based evaluation of the activity or combination of these. The validation studies on Neurosurgery Education and Training School-Skills Assessment Scale (NETS-SAS) identifies the independent parameters of neurosurgery skills as hand-eye coordination, instrument-tissue manipulation, dexterity, flow of procedure and effectualness \cite{c19}. These parameters can be analyzed by the video-based evaluation systems that monitor the activity and movement of the surgeon's hands or tools. The video recording of the activity also provides an opportunity to validate the evaluation using subjective methods.\par 
The video based automatic assessment system can be of two types; offline evaluation and online evaluation. Offline evaluation systems acquire the activity video at reasonable rate and stores the video stream for further analysis. The online evaluation system uses the frame-by-frame analysis, that simultaneously evaluate the activity and also stores it for future reference. \par
Neuro-Endo-Trainer was a box trainer developed for providing skills training for endo-nasal transsphenoidal surgery (ENTS). It was a pick-and-place task trainer that provides the training for basic fundamental skills using standard variable angled neuro-endoscopes \cite{c19}. The evaluation method includes video-based offline evaluation using an auxiliary camera mounted at the top of the box \cite{c*}. The existing method of training on Neuro-Endo-Trainer involves the pick and place of one of the six rings in a predefined pattern under the assistance of technical personnel. The activity performed is sub-divided into sub-activity based on the state of the tool and the rings. The sub-activity can be \emph{``stationary''}, \emph{``picking''} or \emph{``moving''}. The state machine is determined using video processing that includes the tooltip tracking, background segmentation, and ring segmentation. The definition of state machine with the heuristics determined from the video, causes uncertainty and requires a robust task definition system. Therefore, the hardware of the Neuro-Endo-Trainer was augmented with automatic LED-based task definition to determine the state machine. We have developed a stand-alone training system with Neuro-Endo-Trainer to provide online assessment and real-time feedback and defined it as Neuro-Endo-Trainer-Online Assessment System (NET-OAS). Our online automatic assessment system analyzes the activity frame-by-frame and categorizes it as a sub-activity. The relevant parameters of skills training are identified by statistical analysis of the sub-activity. It provides a warning to the trainee neurosurgeon when they make mistakes and provide a detailed synopsis at the end of the activity. The aim of the current study is to validate the developed NET-OAS to establish the level of skills acquisition after staged practice.   

\section{Background}
The low fidelity box-trainers are widely available for laparoscopic skills training \cite{c12, c36} whereas they are limited for neuro-endoscopy. The evaluation system for these trainers can be based on subjective or objective measures. The objective evaluation includes Likert-scale based direct observation, sensor-based evaluation and computerized video analysis. The webcam based endoscopic endonasal trainer developed by Hirayama et al. studied the effectualness of the training by evaluating the performance on LapSim simulator before and after the training \cite{c6}. Neuro-Endo-Trainer SkullBase-Task-GraspPickPlace developed by Raman et.al was validated using subjective evaluation on different target groups \cite{c19}.\par
The video-based evaluation of the surgical activity includes the tracking of the tooltip or tracking the surgeon's hands. There are evaluation systems that use statistical color based image segmentation and tool tracking to identify the tool position and orientation \cite{c26, c27}. The automated skills evaluation method in minimally invasive laparoscopic surgeries were done by segmenting the task into sub-tasks (Therbligs) and their kinematic analysis \cite{c23}. The feature based tool tracking combined with region-based level set segmentation was used to obtain 3D pose estimation of the instruments and to evaluate the psychomotor skills \cite{c44}. There are methods that capture the activity of the subject and track the hand movements using multiple camera feeds \cite{c25}. Neuro-Endo-Activity-Tracker provided a video-based automatic evaluation using Gaussian Mixture based background subtraction and tracking of the tooltip using Tracking-Learning-Detection algorithm \cite{c*}.
\section{Methodology}
NET-OAS consists of low-cost endoscopic system of USB based endoscopic camera that captures the video at 25 fps, variable-angled scopes $(0^0, 30^0, 45^0)$, LED-based light source, Neuro-Endo-Trainer SkullBase-Task-GraspPickPlace box-trainer mounted with GigE based auxiliary camera, and online evaluation software.
\subsection{NET-OAS hardware design}	
The online evaluation system consists of a LED-based task indication method which helps the user to place the ring on the illuminated peg without the assistance of any technician. The peg was illuminated to provide the indication for placement of the ring. The peg plate was printed in two parts: front part of the peg was printed using transparent material by Stereolithography (SLA) technique and back part of the plate was printed using fused deposition modeling (FDA) technique and then both parts were joined using a strong adhesive. The LED array was connected to control circuit using a multiplexer (CD74HC4067). 
\begin{figure}[tbp]
\centering
\includegraphics[width=1\hsize]{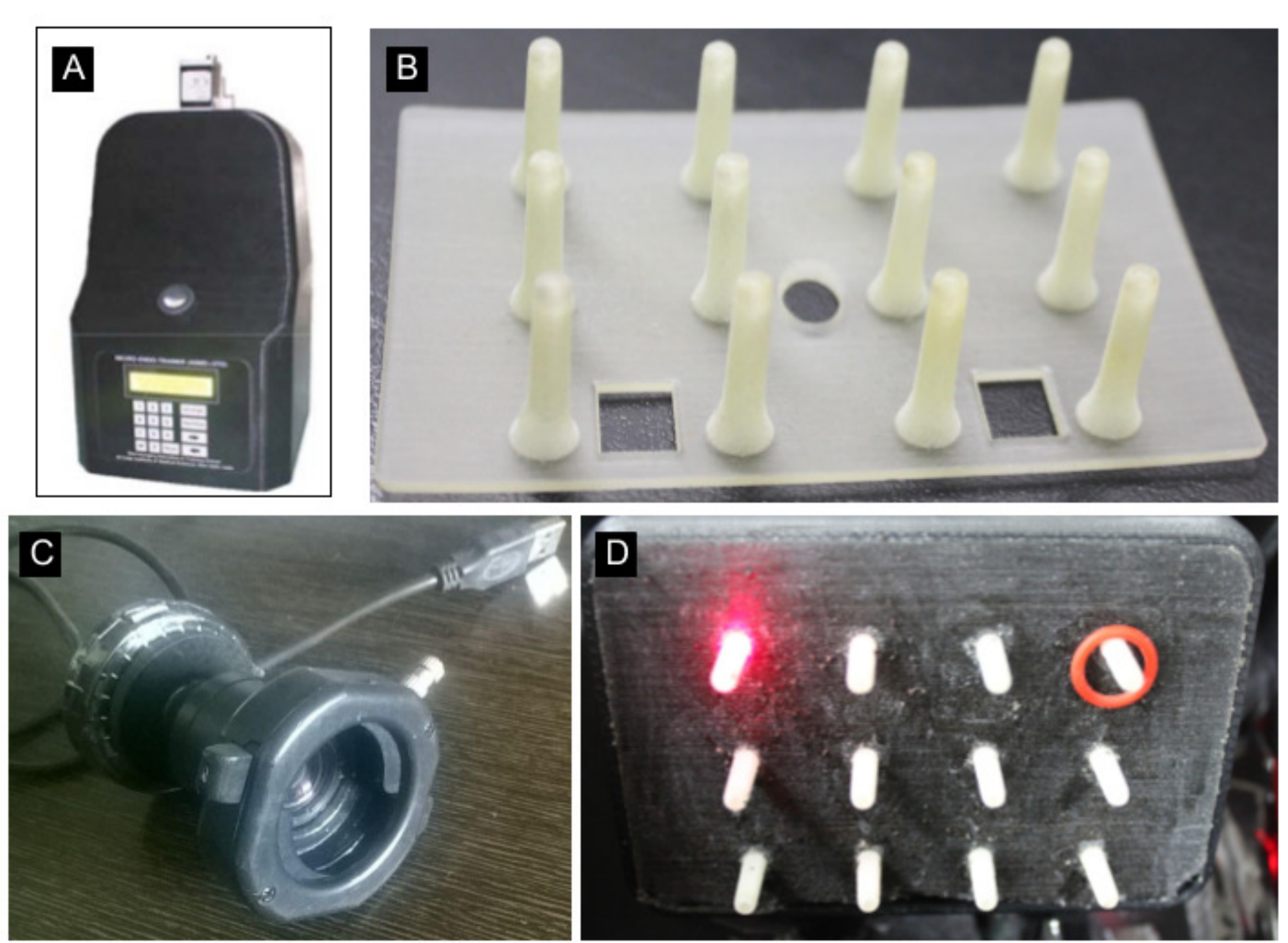}
\caption{A. Neuro-Endo-Trainer SkullBase-Task-GraspPickPlace box-trainer mounted with GigE based auxiliary camera, B. Transparent front-part of the peg plate, C. USB camera with endoscope coupler, D. Peg plate with LED}\label{p1}
\end{figure}
The control circuit consists of ATMEGA328 8 bit micro-controller for the processing, MCP23017 I/O port expander for I/O expansion, 16x2 LCD for display, keypad to provide input, servo motor to control the peg plate and FT232RL serial communication chip to communicate with the PC using serial communication protocol.
There are two cameras in the setup; Low-cost USB based endoscopic camera for the visualization of the site that captures feed at 25fps and GigE based auxiliary camera (Basler ACE) capturing at 50 fps for the online evaluation and real-time feedback. The hardware components of NET-OAS is shown in Fig. \ref{p1}.
\subsection{NET-OAS software design}

\begin{figure*}[tbp]
\centering
\includegraphics[width=1.0\hsize]{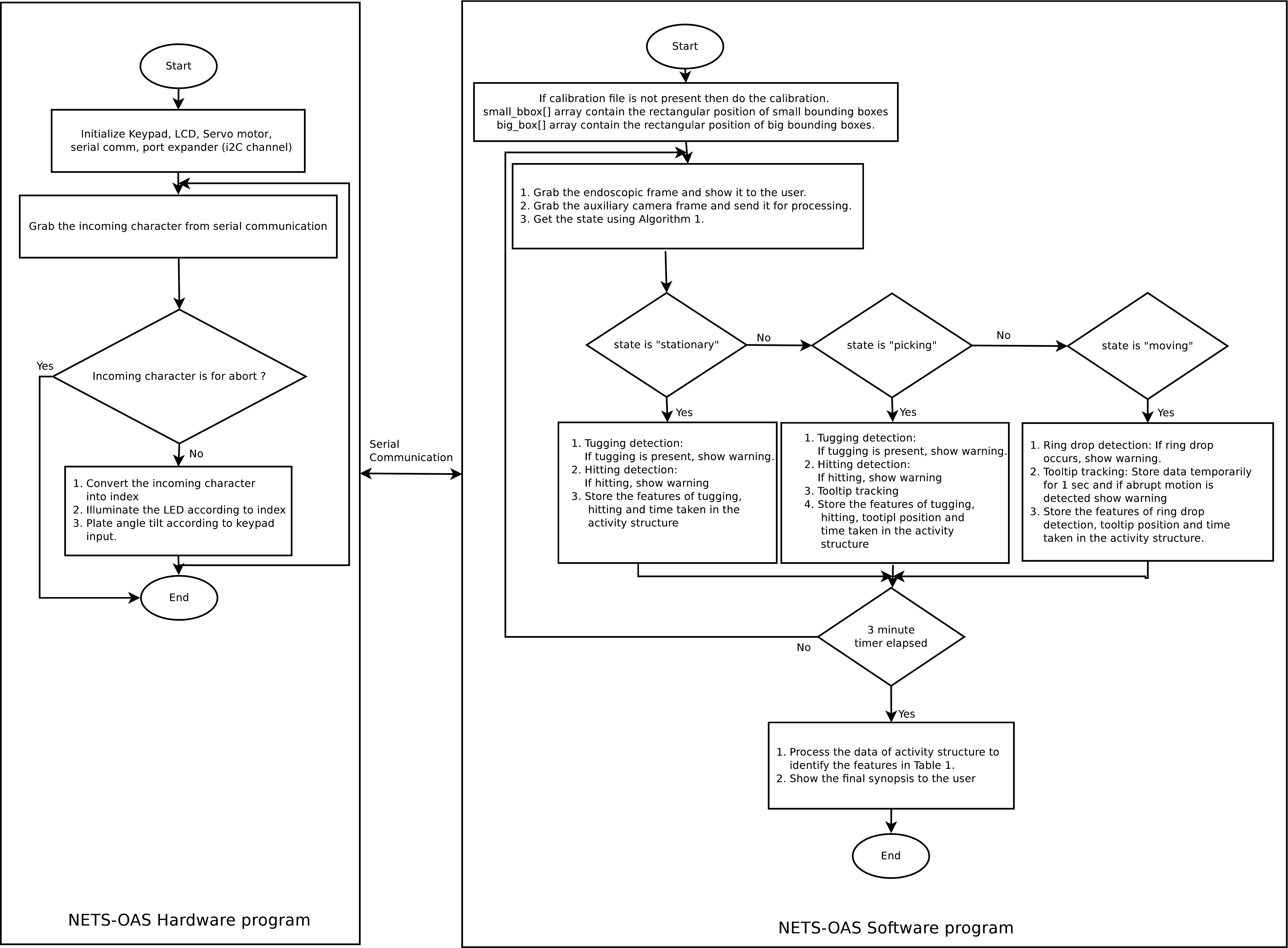}
\caption{Flow Diagram of NET-OAS}\label{p2}
\end{figure*}

The software system of NET-OAS uses a multi-threaded program that processes the two camera streams independently, which maintains the real-time requirement of the system. The complete flow diagram of the NET-OAS is shown in Fig.\ref{p2} and its user interface is shown in Fig.\ref{p3}. It shows endoscopic and auxiliary streams, options to add the user to the database, configure serial port parameters, select the level of training and option to perform calibration if required. When the user hit the Run button, a new window opens the endoscopic stream with screen display of real-time feedback. After the completion of the activity, the results are shown to the user.\par
The main components of the software system are as follows:
\begin{figure}[tbp]
\centering
\includegraphics[width=1\hsize]{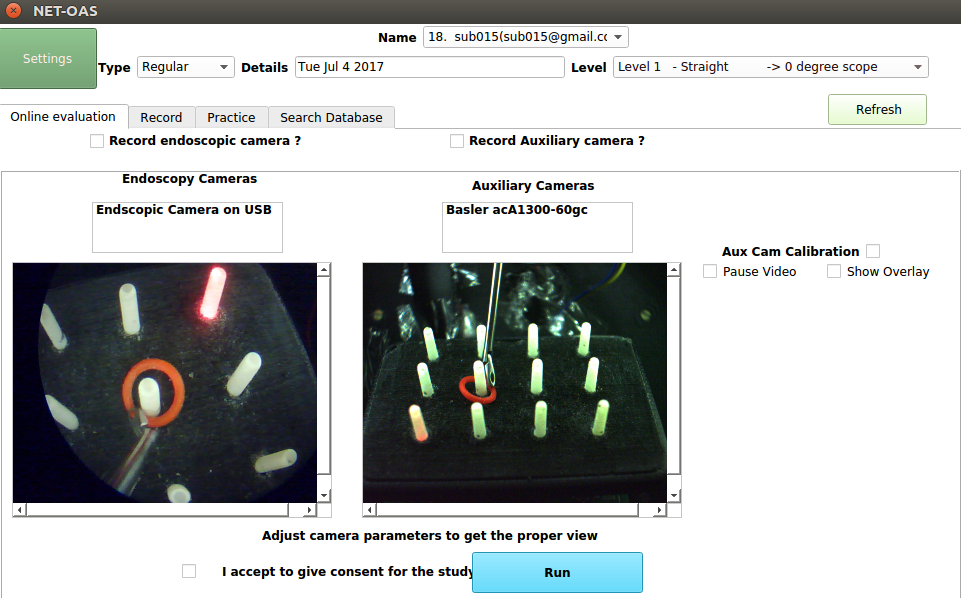}
\caption{User interface of NET-OAS}\label{p3}
\end{figure}

\begin{algorithm}
  \caption{Determine the state machine}\label{state-machine}
  \begin{algorithmic}
   \State $Read\ the\ calibration\ file; Initialize\ small\_bbox[], \ big\_bbox[],$
   \State $\ thresh\_stationary,\ thresh\_picking,\ thresh\_moving$
   \State $ring\_id \leftarrow -1$; $led\_id\leftarrow -1$;
   \State $index\_set \leftarrow true$; $status \leftarrow ``stationary''$;
   \Function{get-State}{$image$}
	   \If{$index\_set$}
		   \State $index\_set \leftarrow false$
		      \For{\texttt{k = 0; k < 12; k++}}
				\State $seg\_image\leftarrow ringSegmentation(image)$;
				\State $sum\_pixels \leftarrow seg\_image[small\_bbox[k]]$;
				\If{$sum\_pixels \le thresh\_picking$}
					\State $ring\_id \leftarrow k+1$;
					\State $break$;
				\EndIf
			  \EndFor
	   \EndIf
	   \State $led\_id\leftarrow random(1-12)$;
	   \State $litLED(led\_id)$
       \State $seg\_image\leftarrow ringSegmentation(image)$;
       \State $s\_old\_small \leftarrow seg\_image[small\_bbox[ring\_id]]$;
       \State $s\_old\_big \leftarrow seg\_image[small\_bbox[ring\_id]]$;
       \State $s\_current\_small \leftarrow seg\_image[small\_bbox[led\_id]]$;
       \If{$status == ``stationary''$}
       		\If{$s\_old\_small \ge thresh\_stationary$}
       			\State $status \leftarrow ``stationary''$;
	        \Else
       			\State $status \leftarrow ``picking''$;
       		\EndIf
       \ElsIf {$status == ``picking''$}
	       \If{$s\_old\_big \ge thresh\_picking$}
    		   \State $status \leftarrow ``picking''$;
    	   \Else
	    	   \State $status \leftarrow ``moving''$;
	       \EndIf
	   \ElsIf{$status == ``moving''$}
		   \If{$s\_current\_small \ge thresh\_moving$}
		   		\State $status \leftarrow ``stationary''$;
		        \State $ring\_id \leftarrow led\_id$;
		        \State $led\_id\leftarrow random(1-12)$;
		        \State $litLED(led\_id)$
       	   \Else
		       \State $status \leftarrow ``moving''$;
       	   \EndIf    
	   \EndIf
	   \State $return \  status$
   \EndFunction
  \end{algorithmic}
\end{algorithm}

\subsubsection {Calibration setup}
\begin{figure}[tbp]
\centering
\includegraphics[width=0.8\hsize]{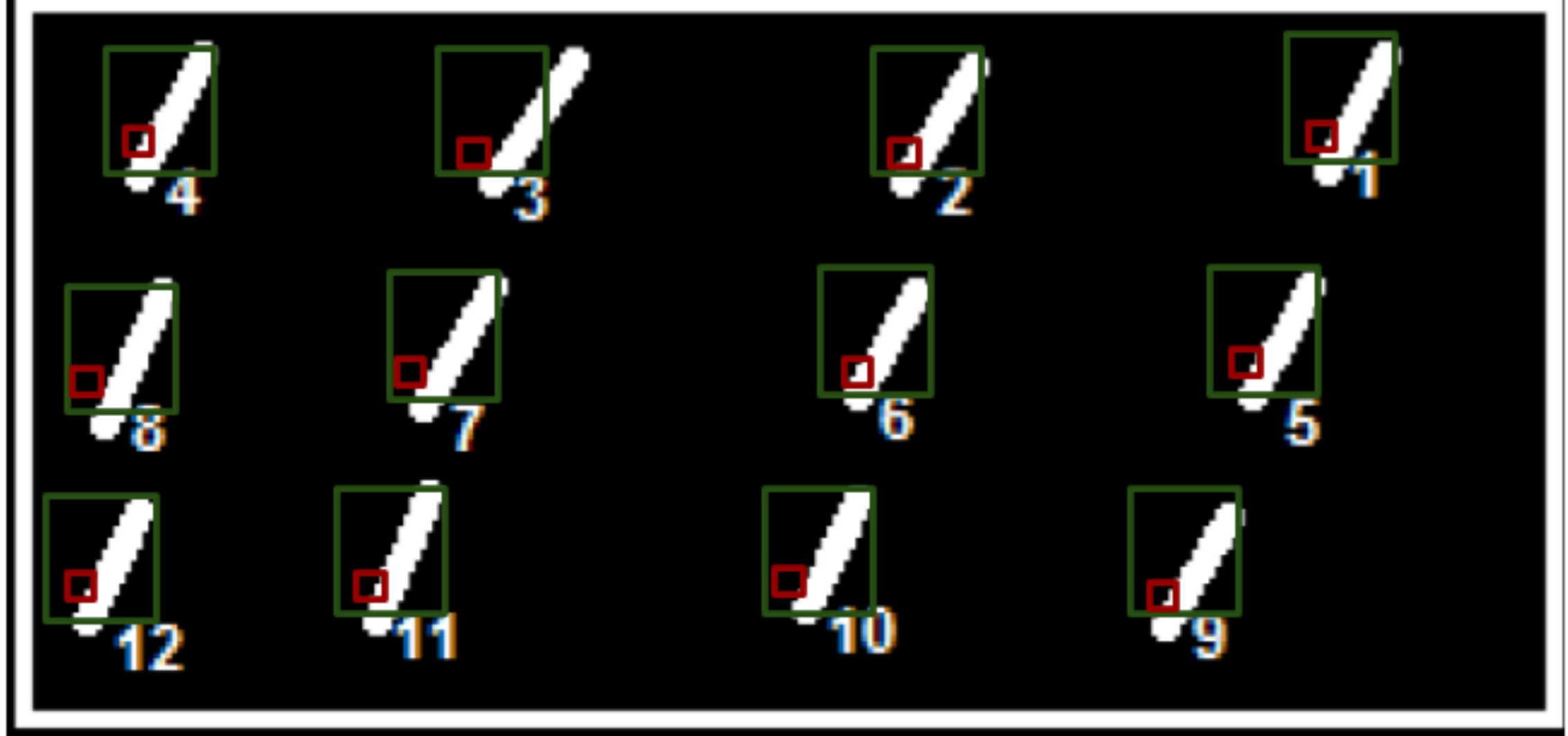}
\caption{Bounding box of pegs}\label{p4}
\end{figure}

One-time calibration involves peg-segmentation, ring segmentation, and tooltip bounding box selection and storing the parameters in the calibration file. The $small\_bbox[]$ contains the rectangular location of small bounding boxes, $big\_bbox[]$ contains the location of big bounding boxes as shown in  Fig.\ref{p4}. These arrays are used to determine the state machine explained in Algorithm~\ref{state-machine}. When the software starts, it loads the parameters from the calibration file otherwise prompt the user to perform the calibration. 

\subsubsection {State machine estimation}
\begin{figure}[tbp]
\centering
\includegraphics[width=0.8\hsize]{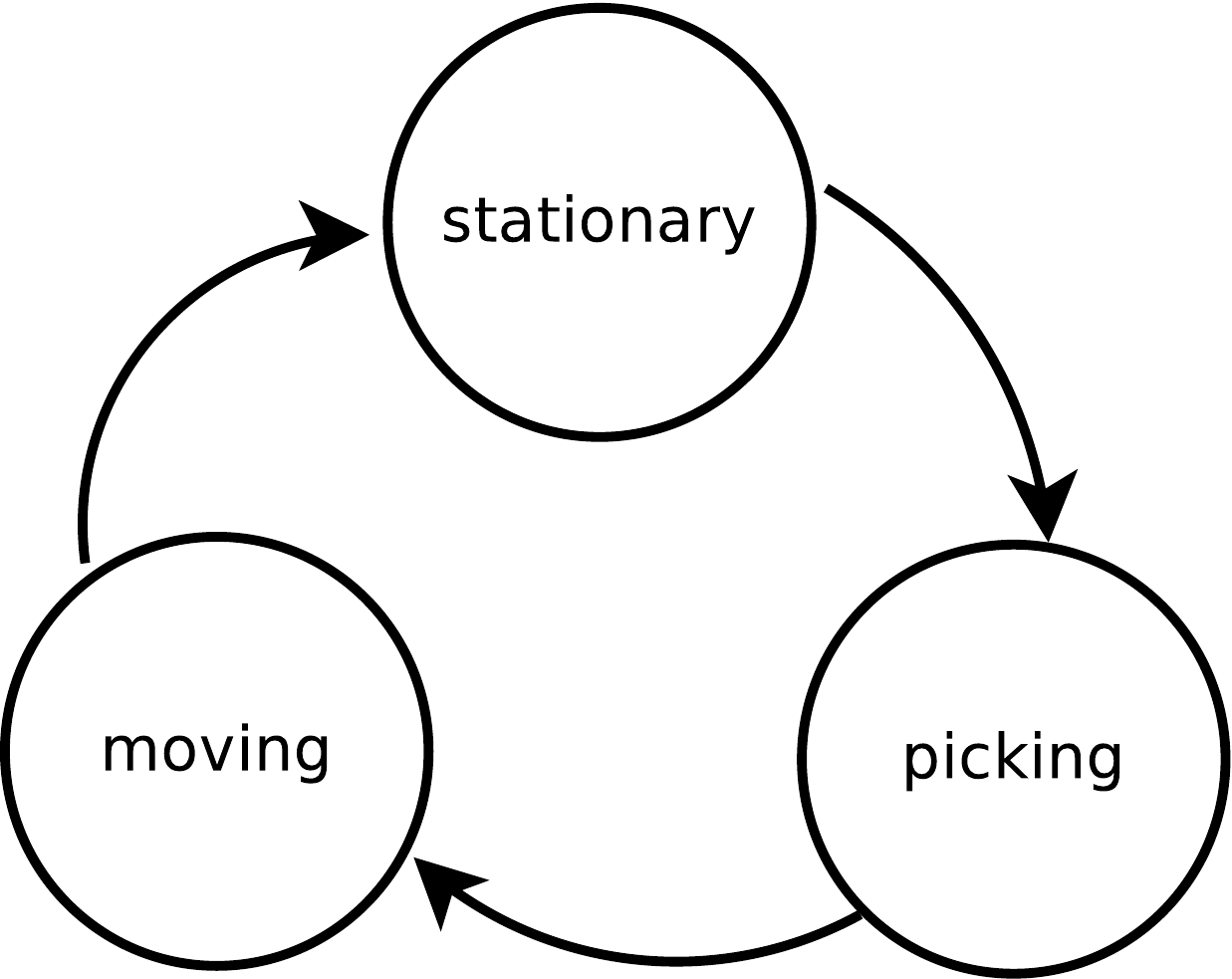}
\caption{State-machine}\label{p5}
\end{figure}

The activity on the NET-OAS in a particular frame can be any of the following sub-activity: \emph{``stationary''}, \emph{``picking''} or \emph{``moving''}. The state machine is initialized with the \emph{``stationary''} state and the states are updated according to the movement of the ring. The \emph{``stationary''} state is defined when the ring is stationary and the tool is present/absent. The \emph{``picking''} state is defined when the tool is near the peg trying to grab the ring till the ring moves out of the peg. The \emph{``moving''} state is defined when the ring has moved out of the peg until it is placed on the illuminated destination peg. Once the ring has been placed on the peg, the ring segmentation output in the bounding box changes and another peg is illuminated randomly. The state machine is unidirectional and cyclic as shown in Fig.\ref{p5}. The algorithm for state machine estimation is explained in Algorithm~\ref{state-machine}. Function $ringSegmentation(image)$ perform the ring segmentation on the input frame and $litLED(int\ number)$ function illuminate the corresponding peg given in its argument.

\subsubsection {Tracking Algorithm}
\begin{figure}[tbp]
\centering
\includegraphics[width=1.0\hsize]{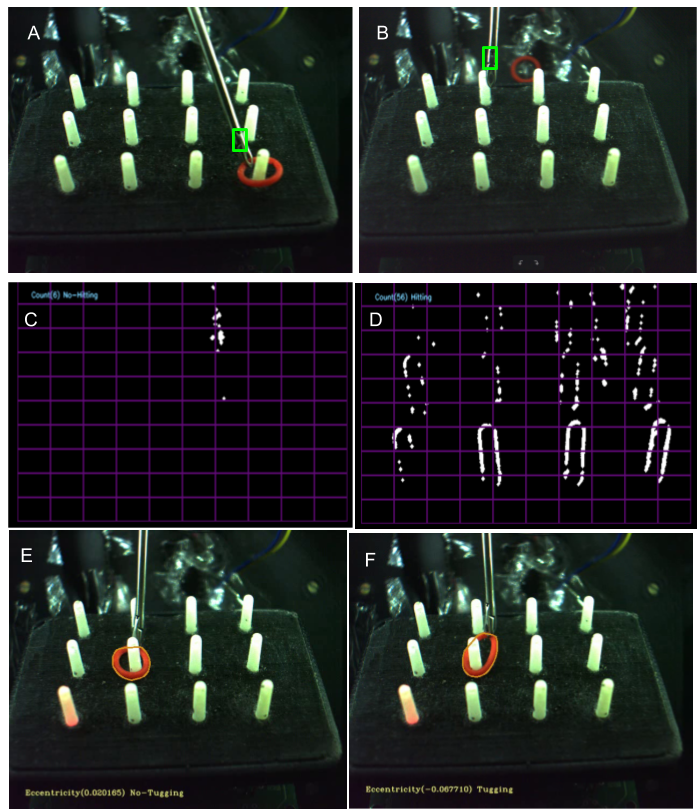}
\caption{Auxiliary camera frame analysis showing: A. Tracking of the tool using TLD algorithm, B. Ring drop determined by the distance between tool-tip and ring segmentation.
C. No Hitting D. Hitting determined by counting the subwindows having significant number of contours, E. No- Tugging F. Tugging determined by eccentricity analysis of the ring contour}
\label{pH}
\end{figure}

Tracking-Learning-Detection (TLD) algorithm is used to track the tooltip. TLD initializes from the bounding box and tracking model, retrieved from the calibration file. It is a robust tracking algorithm which tracks the tooltip under blurred conditions and various transformations. The tracking is based on median flow tracker which track the tooltip frame-to-frame and measure the tracking error using efficiency of backtracking. The detection thread is a 3-stage sliding window cascaded classifier, which consists of variance filter, random forest, and nearest neighbor classifier. At the end of the 3rd stage, it provides a set of windows that localizes the appearance of the tool tip. It predicts the next location of the tool tip having the minimum error in tracking or detection stage. The remaining set of appearances is fed to the  negative class for better generalization of the tool tip model. Tracking of the tool using TLD algorithm is shown in Fig.\ref{pH} A.\cite{c46}.
\subsubsection {Ring Drop Detection}
The dropping of the ring is determined in the \emph{``moving''} state if distance between the tool tip bounding box (determined by TLD) and the $ringSegmentation(image)$ is more than a predefined threshold. Fig.\ref{pH} B shows the image of the ring drop condition.

\subsubsection {Hitting Detection}
The hitting of the peg board happens due to poor depth perception of the user. The hitting is detected using image analysis of the successive frames. The difference image is divided into 10x10 grids and hitting is recorded by identifying the number of grids that shows significant movement. The hitting threshold is set experimentally and the Fig.\ref{pH} C shows the case of no hitting and Fig.\ref{pH} D shows a hitting instance output.

\subsubsection {Tugging detection}
The tugging is detected by analyzing the deformation of the ring in the \emph{``stationary''} and \emph{``picking''} state. The ring is segmented based on the hue value obtained from the calibration file. Due to the overlapping of the tool or peg, $ringSegmentation(image)$ results in two or more contours. The contour with maximum size and the nearest contours are determined and combined. The $$eccentricity = \frac{\mu_2,_0 + \mu_0,_2 + \sqrt{(\mu_2,_0 - \mu_0,_2)^2 + 4(\mu_1,_1)^2}}{\mu_2,_0 + \mu_0,_2 - \sqrt{(\mu_2,_0 - \mu_0,_2)^2 + 4(\mu_1,_1)^2}}$$ 
$$\mu_i,_j = \sum_{x} \sum_{y} x^iy^jI(x,y) $$
value of the combined contour is sufficient to determine the deformation of the ring in case of tugging. The eccentricity threshold corresponding to tugging is set experimentally.

\subsubsection {Tracking data analysis}
Tracking data analysis is done to identify motion smoothness and sudden jerk of the tool tip motion in the \emph{``moving''} state. Smoothness of the path is measured by taking the standard deviation of the first derivative of the tracking data, Arc length of the path is measured by counting number of pixels of the tracking data in the \emph{``moving''} state. Curvature at each point of tracking data is computed using $$\kappa = \frac{ \lvert(\frac{\partial x}{\partial t} * \frac{\partial^2 y}{\partial t^2}) - (\frac{\partial y}{\partial t} * \frac{\partial^2 x}{\partial t^2})\rvert }{(\frac{\partial x}{\partial t}^2 + \frac{\partial y}{\partial t}^2)^\frac{3}{2}}$$

\subsubsection {Real time feedback}
At each frame, the algorithm identifies the current state and provide real time feedback for hitting, tugging and ring drop. Motion smoothness feedback is provided after processing frames of last 1 second. The output is displayed on the endoscopic screen to warn the user. This helps the user to learn and correct the mistakes accordingly.

\subsubsection {Feature Extraction and final synopsis}
The activity data structure stores the current sub-activity (\emph{``stationary''}, \emph{``picking''} or \emph{``moving''}) and its related parameters as shown in Table 1. At the end of the activity, the data is processed to give the final synopsis to the user.
\begin{table}[tbp]
\caption{Selected features for NET-OAS}\label{table}
\centering
\begin{tabular}{|@{\vrule width0ptheight11pt\enspace} p{4cm}|p{4cm}|}\hline
\bf Measure from NETS-SAS &\bf Selected objective measure for NET-OAS
\\\hline
\multirow{2}{*}{\bf \emph{Grasper tissue manipulation
}} & Average time taken to grasp\\\cline{2-2}
						  & Number of tugging events						  
\bf \\\hline
\multirow{2}{*}{\bf \emph{Eye-hand coordination}} & Number of hitting events \\\cline{2-2}
						               & Intensity with which hitting happened
\\\hline
\multirow{4}{*}{\bf \emph{Dexterity}} & Time taken for moving ring from one peg to another  \\\cline{2-2}
						   & Average number of moves			  \\\cline{2-2}			   						   
						   & Smoothness of the path  \\\cline{2-2}
						   & Arc length of the path
\\\hline
\bf \emph{Instrument tissue manipulation} & Number of times curvature value exceeded threshold
\\\hline
\bf \emph{Effectualness} & Number of times ring dropped
\\\hline
\end{tabular}
\end{table}

\section{Experimentation and Results}
\begin{figure}[tbp]
\centering
\includegraphics[width=1.0\hsize]{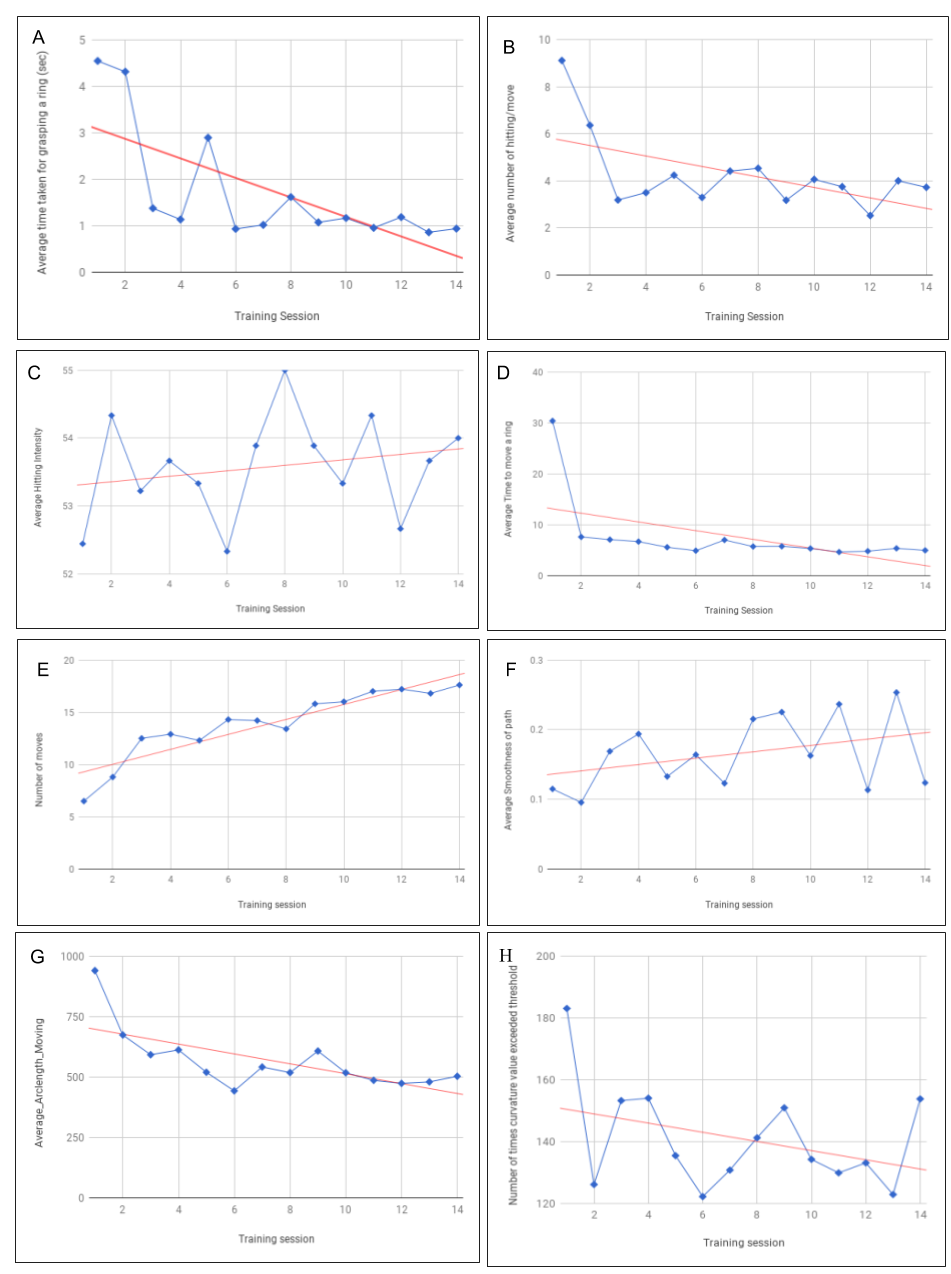}
\caption{Validation study results: Horizontal axes is the training session, blue marker shows the data point and red line shows the trend-line: A. Average time of grasping the ring, B. Average number of hitting, C. Average hitting intensity D. Average time to move a ring, E. Total number of rings placed F. Average smoothness of the tool tip in \emph{``moving''} state, G. Average Arc length of the tool tip in \emph{``moving''} state, H. Number of times curvature exceeded the threshold value or sudden jerk.}
\label{pT}
\end{figure}

A group of 15 novices participated in the study of validation of NET-OAS, who were students from a technical university without any medical training. The demo video demonstrating the good and bad endoscopy practice on Neuro-Endo-Trainer was shown before the practice session. There was a pre-test followed by two sessions and a post-test. The pre-test and post-test included the most difficult task level of $45^0$ scope with right tilt plate. Each activity was programmed to be of 3 minutes duration. The first session consisted of practice using $0^0$ and $30^0$ scopes and with straight, left and right tilts of the plate. The second session was conducted three days later and consisted of practice using $30^0$ and $45^0$ scopes and with straight, left and right tilts of the plate.
Fig. \ref{pT} shows the graph of objective measure for NET-OAS w.r.t training session. The noticeable changes were the increased average number of moves and average smoothness of the path. There were 
decreased number and hitting instances, grasping time, average arc length and sudden jerk motion. The self-assessment feedback obtained from the user also shows that the training session on the NET-OAS made them acquainted with the system.  \par

\subsubsection {Machine learning for validation study}
For the validation study, activity data obtained from 15 novices (pre-test, post-test, 1st trial of session 1 and last trial of session 2) was considered. Pre-test data was considered as \emph{`class novice'} and post-test data was considered as \emph{`class-improved'}. The SVM classifier was trained with 11-dimensional feature vector of these classes. For testing, 1st trial of session 1 was considered as \emph{`class novice'} and the last trial of session 2 was considered as \emph{`class improved'}.
The SVM classifier on the testing data classifies feature set of the 1st trial as \emph{'class novice'} and the last trial of session 2 as \emph{'class improved'} with the accuracy of $88 \%$. \par

\begin{figure}[tbp]
\centering
\includegraphics[width=1.0\hsize]{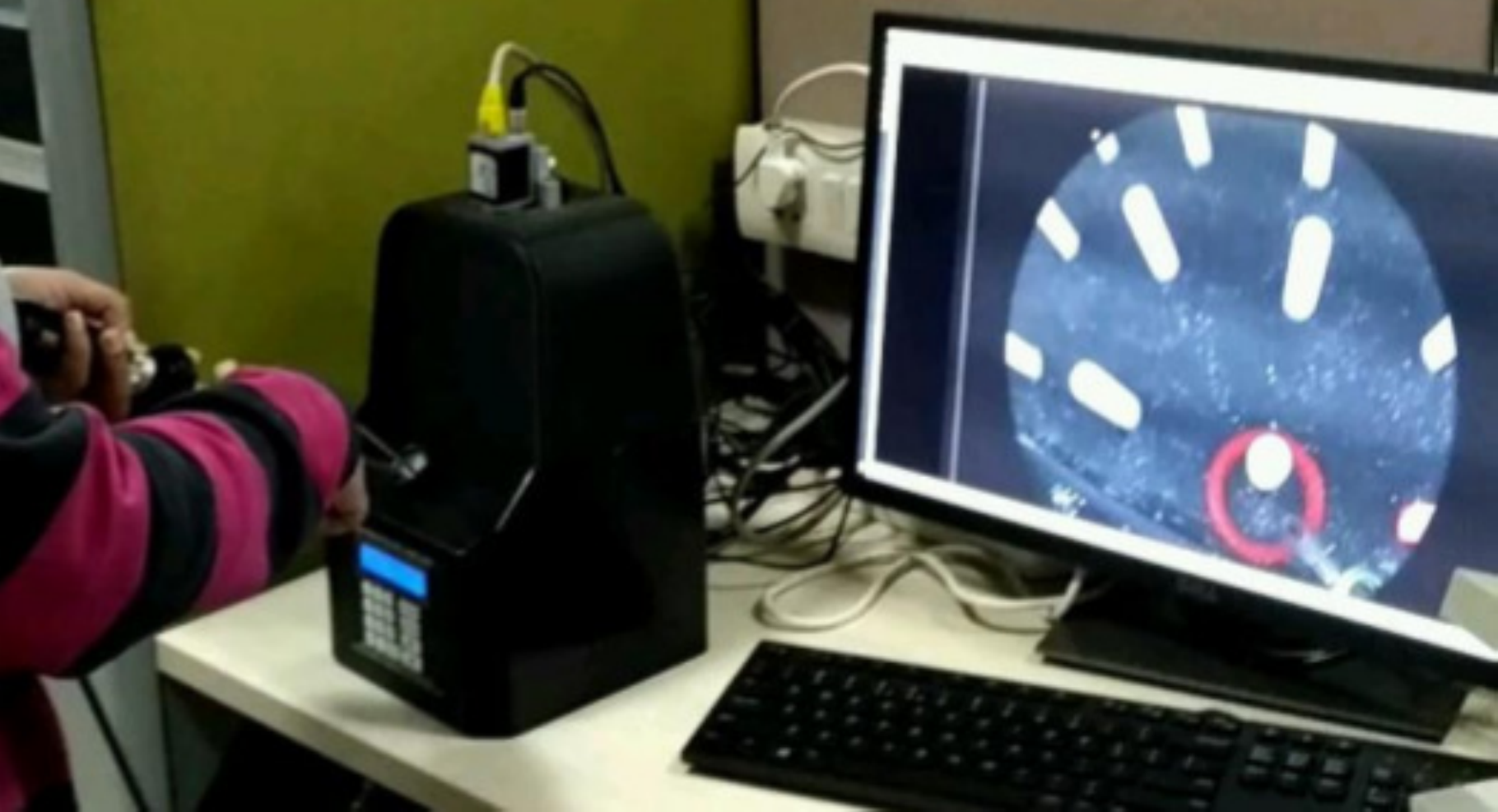}
\caption{Training on the NET-OAS}\label{p6}
\end{figure}

\begin{figure}[tbp]
\centering
\includegraphics[width=1.0\hsize]{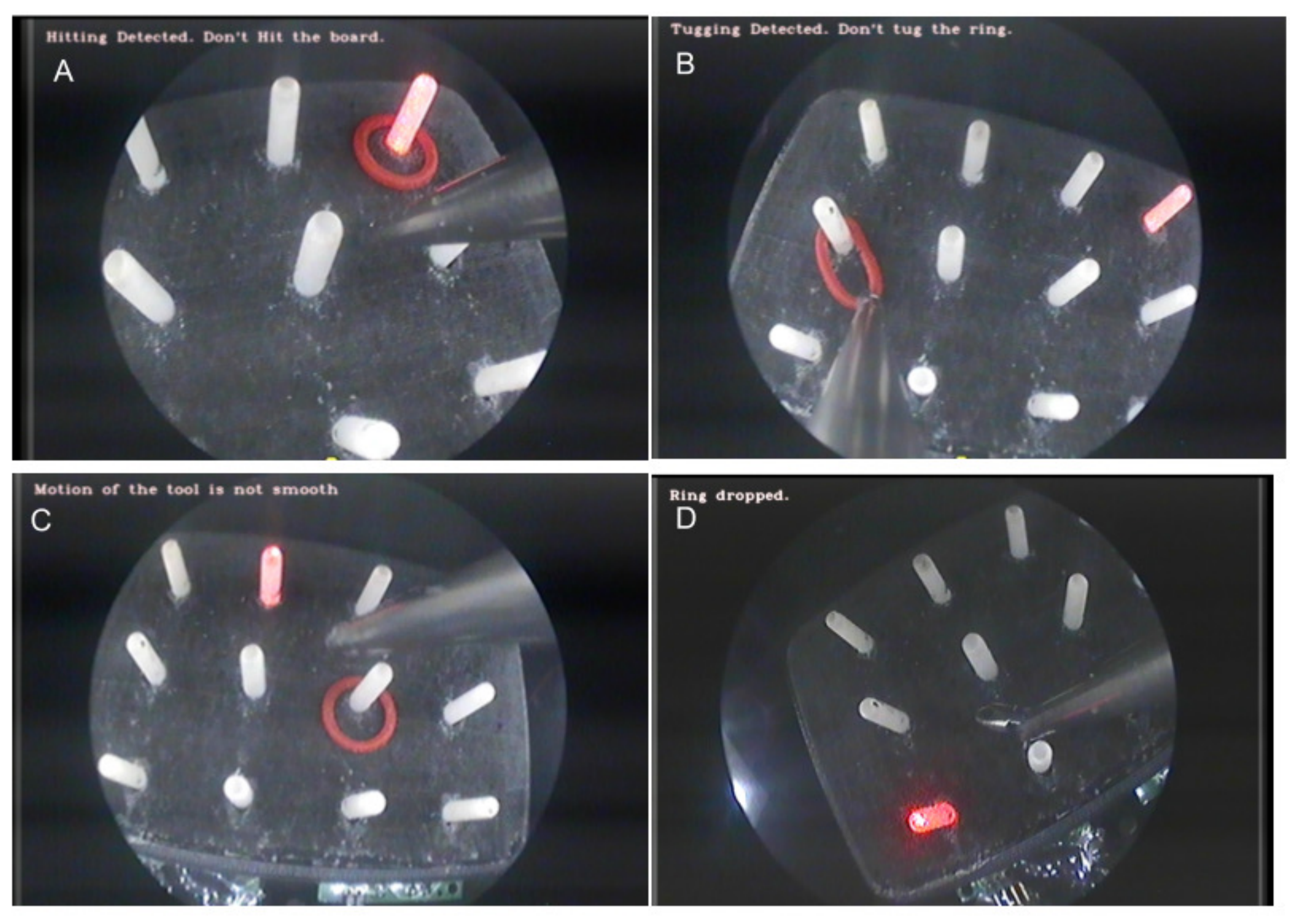}
\caption{Real-time feedback to trainee A) Hitting  B) Tugging  C) Motion smoothness D) Ring Drop}\label{p7}
\end{figure}

The practice session example on the NET-OAS and the real-time feedback provided to the trainee while performing the activity is as shown in Fig. \ref{p6} and Fig. \ref{p7} respectively.

\section{Discussion}
The improvements of NET-OAS as compared to the earlier version include: a complete standalone system, automatic task definition using LED array and serial communication with the hardware, tugging detection algorithm, and ring drop detection. The study used the auxiliary camera for the evaluation of the activity and has not used the endoscopic feed for evaluation. \par The main objective of the study was to validate the NET-OAS on completely novice participants to identify whether there is any improvement in skills acquisition. The results show that after stipulated training on the NET-OAS, the participant improved his/her skills on manipulating the endoscope and tool irrespective of their background. The study can be extended to the intermediate trainee neurosurgeons and experts.
%

\section*{Acknowledgment}
We would like to thank all the participants, research scholars of Indian Institute of Technology Delhi who took part in the study, and the team of Neurosurgery Education and Training School for their support. This work is supported by Department of Health Research, Ministry of  Health and Family Welfare, Govt. of India Project Code No: GIA/3/2014-DHR, Department of Science and Technology (DST), Ministry of Science and Technology, Govt. of India Project Code No: SR/FST/LSII-029/2012.



\end{document}